\documentclass[letterpaper, 10pt, conference]{ieeeconf}      

\usepackage[utf8]{inputenc}
\usepackage[T1]{fontenc} 
\usepackage{graphicx}
\usepackage{latexsym}
\usepackage{color}
\usepackage{mathtools}
\usepackage{amssymb}
\usepackage{amsfonts}

\usepackage{xcolor}
\usepackage{textpos}
\usepackage{setspace}
\usepackage[bookmarks=true]{hyperref}
\usepackage[all]{hypcap}

\hypersetup{
    colorlinks,
    linkcolor={red!50!black},
    citecolor={blue!50!black},
    urlcolor={blue!80!black}
}

\makeatletter
\def\input@path{{../graphs/}}
\makeatother
\graphicspath{{../graphs/}}

\IEEEoverridecommandlockouts                              
\newcommand{\vect}[1]{\boldsymbol{#1}}

\title{\LARGE \bf
Probabilistic Depth Image Registration incorporating Nonvisual Information
}
\author{Manuel Wüthrich, Peter Pastor, Ludovic Righetti, Aude Billard, Stefan Schaal
\thanks{This work has been done at University of Southern California (USC) and has been assigned by École Polytechnique Fédérale de Lausanne (EPFL)}
\thanks{M. Wuthrich is with the Faculty of Micro Engineering, EPFL
        {\tt\small manuel.wuthrich@gmail.com}}%
\thanks{P. Pastor, L. Righetti and S. Schaal are with the Computational Learning and Motor Control Lab (CLMC), USC}%
\thanks{A. Billard is with the Learning Algorithms and Systems Laboratory (LASA), EPFL}%
}

\begin{document}
\maketitle
\thispagestyle{empty}
\pagestyle{empty}

\begin{abstract}
In this paper, we derive a probabilistic registration algorithm for object modeling and tracking.
In many robotics applications, such as manipulation tasks, nonvisual information about 
the movement of the object is available, which we will combine with the visual information.
Furthermore we do not only consider observations of the object, but we also take space into account which has been observed to not be part of the object.
Furthermore we are computing a posterior distribution
over the relative alignment and not a point estimate as typically done in 
for example {\em Iterative Closest Point\/} (ICP). To our knowledge no existing algorithm meets these three
conditions and we thus derive a novel registration algorithm in a
Bayesian framework.
Experimental results suggest that the proposed methods perform favorably in comparison to PCL \cite{rusu11} implementations of feature mapping and ICP, especially if nonvisual information is available.
\end{abstract}

\begin{textblock*}{100mm}(.\textwidth,-10.5cm)
 \begin{spacing}{0.8}
 {\fontsize{8pt}{2pt}\selectfont \sffamily
\noindent 2012 IEEE International Conference on\\
Robotics and Automation (ICRA)\\
May 14-18, 2012. Saint Paul, USA}
\end{spacing}
\end{textblock*}%

\section{INTRODUCTION}
In this paper we will focus on the scenario
where the camera is fixed and only the object is
manipulated. While the object is being moved, a 3D camera gathers depth
images of the object in different orientations and positions. Let us denote two such images as image $A$ and image $B$. The core problem considered in this paper is to estimate the rigid body transformation $\vect{T}$ the object has undergone between the acquisitions of these two images. Segmentation is not the focus of this work, we employ existing algorithms \cite{rusu11} to determine whether a pixel in the depth image belongs to the object or to the background.

A great deal of work has been done in this research area in the past
years. In \cite{cui10} an algorithm is presented which creates 3D
models of objects while the camera or the object is moved. However, the point 
clouds have to be approximately aligned initially and the 
model is created off line by optimizing the alignment of all images simultaneously.
Our method is more general in the sense that point clouds do not have to be approximately aligned. 
However, task-specific assumptions like that can be introduced to significantly 
reduce the computational time for finding the optimal alignment.   

A lot of very promising work, such as \cite{krainin11,li09}, has been
published in the last years about scanning objects while they are being held
by the robot. We however want to treat a more general case where we do
not assume that the object is already grasped or can be grasped in a 
straightforward manner.

In \cite{rusu09} models are constructed by mapping shape primitives to
the point clouds with promising results. In this work however we try
to make as few assumptions as possible about the shape of the object
and thus exclude the use of models or shape primitives.

Among the most popular algorithms that tackle the registration problem
are Iterative Closest Point (ICP) and feature mapping algorithms
and combinations of both 
\cite{dai11,liu02,rusu08,fukai11,rusu09_fpfh}. 
We will compare the proposed method with these two approaches.

ICP has been proven to converge to a local minima \cite{ICP:92}.  In the scenario considered in this paper, an object can move very fast and therefore, point clouds of two subsequent images are not necessarily approximately aligned. This problem is usually tackled by initially aligning point clouds using a feature mapping algorithm \cite{dai11,rusu08,fukai11}. These methods perform well, if different parts of the object can easily be distinguished. For objects with a homogeneous texture, color or local shape, feature matching can be problematic. Furthermore if the quality of
the features degrades with the quality of the point cloud, noisy data can cause problems.

Often in robotics there is a great deal of nonvisual information about the
transformation of the object available. In our scenario, this information can for example be that an object is pushed on a table and the movement will therefore be in a plane.  If it is held by a robot, we approximately know how the object will move. This kind of information can certainly be incorporated in ICP and feature mapping algorithms, but they are not originally designed to do so.

ICP and feature mapping algorithms commonly optimize a cost function that is only dependent on the relative alignment between two point clouds. In our proposed method, we take into account the space which has been observed to not contain any part of the object. Our results suggest that taking this information into account leads to more robust registration results. 
Introducing visibility constraints has previously been shown to help in estimating the occluded shape of an unknown object \cite{bohg:icra11}.

Finally, feature mapping and ICP algorithms usually return a point
estimate of the transformation and a fitness. It can however be preferable to have a more differentiated estimate of the
transformation in form of a probability distribution over the 6
parameters of the transformation. This allows us to express, for
example, that we are certain about the rotation around axis $x$ but
uncertain about the translation in $y$ etc. In the results section we
will show an example of the use of a probability distribution as
result.

To our knowledge, there is no registration algorithm that combines the three mentioned points:
\begin{enumerate}
\item Cost function based on visibility constraints.
\item Output of a posterior distribution over the estimated object pose change.
\item Straightforward incorporation of task-relevant nonvisual information.
\end{enumerate}
In the next section, we will derive the proposed registration algorithm in a Bayesian framework. In the result section, we show that under certain conditions that are quite common in the scenario of object model learning and tracking, our algorithm outperforms implementations of ICP and feature mapping methods.

\section{DERIVATION}

\subsection{Incorporated Information}
\begin{figure}[h]
 \centering
 \includegraphics[scale=0.25]{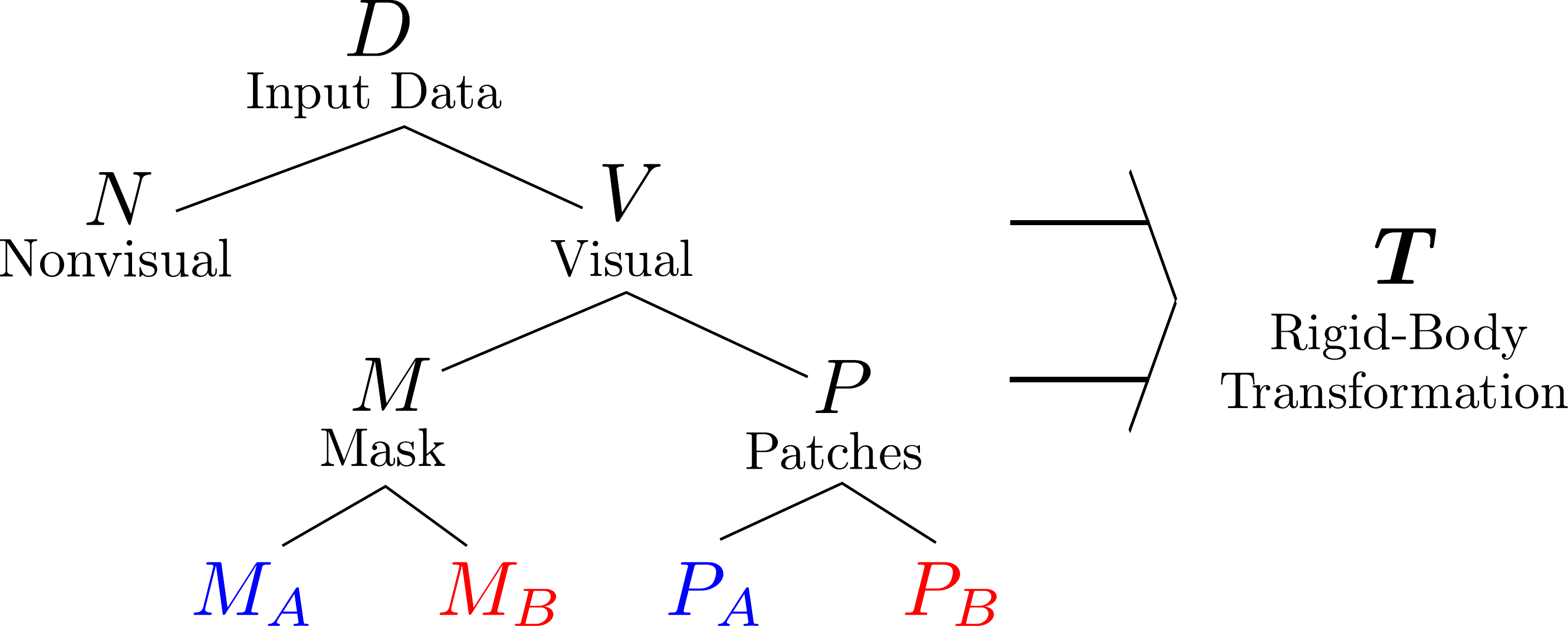}
 \caption{Overview of the variables}
 \label{fig:variables}
\end{figure}

An overview of all the information we will make use of can be seen in Fig.~\ref{fig:variables}. The input data $D$ consists of the visual information $V$ and the nonvisual information $N$.
\subsubsection{Nonvisual Information}
In the context of a robotic manipulation task often a great deal of nonvisual information about the movement of an object is available.
$N$ can contain for example the information that the object will be moved on a table, that the robot has poked it with a certain movement or that the object is being held by the robot, and we thus know how it has moved approximately. \\

\subsubsection{Visual Information}
\begin{figure}[h]
 \centering
 \includegraphics[scale=0.4]{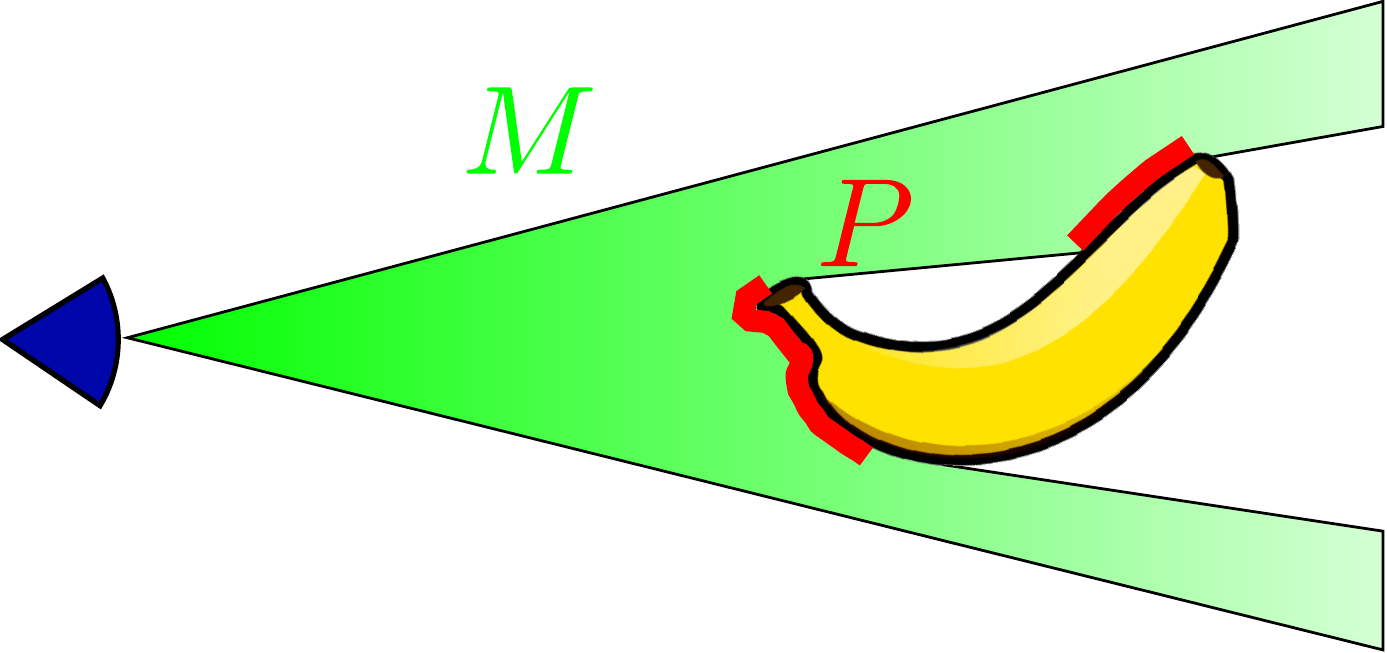}
 \caption{Two types of visual information: Surface patches $P$ and mask $M$.}
 \label{fig:concept}
\end{figure}

We divide the visual information into two types (see Fig.~\ref{fig:concept}). Firstly there are surface patches which are observed by the depth camera, from now on referred to as patches $P$. These patches can be represented as a point cloud and are thus the only information used by ICP and by most feature mapping algorithms.\\
There is however another very important piece of information. No part of the object is inside the green area in Fig.~\ref{fig:concept}, this area defines thus a mask $M$ for the object.\\
We will always register two depth images, $A$ and $B$, at a time, therefore we have of course the masks, $M_A$ and $M_B$, as well as the patches, $P_A$ and $P_B$, from each image (see Fig.~\ref{fig:variables}).

\subsection{Parametrization}
\subsubsection{Coordinate system}
\begin{figure}[h]
 \centering
 \includegraphics[scale=0.6]{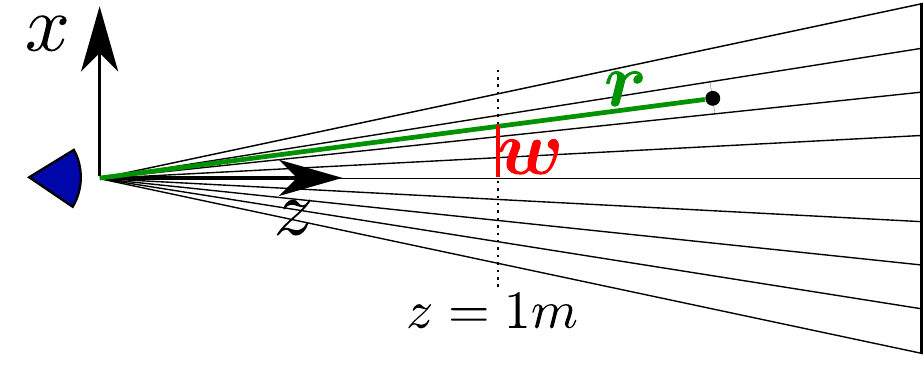}
 \caption{Schematic representation of eight pixels acquired by the
   depth camera. The coordinates $r$ and $w$ are represented, while $h$ would be perpendicular to the image plane.}
 \label{fig:coordinates}
\end{figure}

Given that we will work with depth images we choose a suitable parametrization assuming the pinhole model for the camera. The first two parameters, $w$ and $h$, are chosen to be the projections of a 3D point onto a virtual image plane given a focal length of 1m, see Fig.~\ref{fig:coordinates}.  
%
%
The third parameter $r$ is the depth of the 3D point.
These coordinates will be called ray
coordinates. They are derived
from Cartesian coordinates as follows:
\begin{align}
w &= \frac{x}{z}, ~ h = \frac{y}{z}, ~r = \sqrt{x^2 + y^2 + z^2}
\end{align}
\subsubsection{Rigid body transformation}
The rigid body transformation $\vect{T}$ has six independent parameters $\vect{T}=(T_1,...,T_6)^{\top}$. The parametrization can be chosen to be whatever is convenient for a given application.
\subsection{Measurement Error}
Due to measurement errors in the camera, an observed 3D point $\vect{p}$ will not exactly correspond to the true point $\vect{s}$ on the object surface. As a measurement model $p(\vect{p}|\vect{s})$ we use a normal distribution in ray coordinates.
\begin{align}
 p(\vect{p}|\vect{s}) &= \mathcal{\mathcal{N}}(\vect{s}|\vect{p}, L) \label{eq:emissionprob}
\end{align}
The covariance matrix $L$ is camera specific. The only assumption we make in our derivation is that the covariance matrix is such that  $p(\vect{p}|\vect{s})$ can be reasonably well approximated as being constant within a pixel. This assumption is sensible because the depth camera is not able to distinguish between points within the range of one pixel.\\
Furthermore, assuming that $p(\vect{p})$ and $p(\vect{s})$ are uniform in the range of the depth camera, we have  $p(\vect{s}|\vect{p}) = p(\vect{p}|\vect{s})$.\\

\subsection{Derivation}
Our objective is to express $p(\vect{T}|D)$, the probability distribution over the transformation $\vect{T}$ the object has undergone, given all the available data $D$. Applying Bayes we have%
\begin{align}
p(\vect{T}|D) &= p(\vect{T}|N,P,M)\\
	    &=\frac{ p(M|N,P,\vect{T}) p(\vect{T}|N,P)}{\int p(M|N,P,\vect{T})  p(\vect{T}|N, P)d\vect{T}} \label{eq:pTDD}
\end{align}
In $p(M|N,P,\vect{T})$, given the transformation $\vect{T}$, the mask $M$ does not depend on the nonvisual information $N$ and we thus have
\begin{align}
p(M|N,P,\vect{T}) &=p(M|P,\vect{T})\\
 &= p(M_A, M_B|P_A, P_B, \vect{T})
\end{align}%
We assume $M_A$ and  $M_B$ to be independent because the mask observed in one image does not give us any useful information about the mask observed in the other image.%
\begin{align} p(M|N,P,\vect{T}) &= p(M_A|P_A, P_B, \vect{T}) p(M_B|P_A, P_B, \vect{T}) \end{align}
As $M_A$ and $P_A$ are from the same image, the object $P_A$ will necessarily be respected in the mask $M_A$. $P_A$ does thus not add any information to the first term and can be removed. Similarly, for the second term we can omit $P_B$.%
\begin{align} p(M|N,P,\vect{T})&= p(M_A|P_B, \vect{T}) p(M_B|P_A, \vect{T}) \end{align}
It is reasonable to assume that the priors $p(M|\vect{T})$ and $p(P|\vect{T})$ are uniform because we do not have any prior information about the distribution of the points and the mask. Applying of Bayes' rule, we thus have%
%
\begin{align}
\addtolength{\fboxsep}{5pt} 
p(M|N,P,\vect{T})=k p(P_B| M_A, \vect{T}) p(P_A| M_B, \vect{T})
\end{align}
with k being a constant.

Inserting this result into Eq.~\ref{eq:pTDD} we obtain
\begin{align}
p(\vect{T}|D)  &=\frac{ p(P_B|M_A,\vect{T}) p(P_A|M_B,\vect{T})p(\vect{T}|N, P)}{\int p(P_B|M_A,\vect{T}) p(P_A|M_B,\vect{T})p(\vect{T}|N, P)d\vect{T}}\notag
\end{align}
Finding this distribution is intractable, but for most purposes we do not need the distribution itself, we only use it for evaluating expectations. We thus need to find the expectation of a function $f(\vect{T})$ expressing a property of $\vect{T}$ required for a given application. If $f$ is for example identity ($f(\vect{T}) = \vect{T}$), then $E(f(\vect{T})) = E(\vect{T})$, or if $f(\vect{T}) = (\vect{T} - E(\vect{T})) (\vect{T} - E(\vect{T}))^\top$  then $E(f(\vect{T}))$ is the covariance matrix. The expectation of a function of $\vect{T}$ is  
\begin{align}
	   &E(f(\vect{T}))    =\\
	   &\int \frac{ p(P_B|M_A,\vect{T}) p(P_A|M_B,\vect{T})p(\vect{T}|N, P)}{\int p(P_B|M_A,\vect{T}) p(P_A|M_B,\vect{T})p(\vect{T}|N, P)d\vect{T}}f(\vect{T})d\vect{T} \notag\\
  \Aboxed{ &E(f(\vect{T}))  \approx \sum_{l=1}^L w^{(l)}  f(\vect{T}^{(l)})}\label{eq:sampling}
  \end{align}
  Where the samples $\vect{T}^{(l)}$ are drawn from $p(\vect{T}|N,P)$. The sampling weights $w^{(l)}$ are defined by
  \begin{align}
\Aboxed{&w^{(l)} = \frac{p(P_B| M_A, \vect{T}^{(l)}) p(P_A| M_B, \vect{T}^{(l)})}{\sum_{m=1}^Lp(P_B| M_A, \vect{T}^{(m)}) p(P_A| M_B, \vect{T}^{(m)})} } \label{eq:sampling2}
  \end{align}
We thus have represented  $p(\vect{T}|D)$ by a set of samples $\{\vect{T}^{(l)}\}$ and the corresponding weights $\{w^{(l)}\}$. The samples are drawn from $p(\vect{T}|N,P)$, in other words, we will create a distribution, from which it is possible to sample, taking into account the nonvisual information as well as the observed surface patches. This distribution will be defined independently for a given application, an example is discussed in the results section.\\
The terms $p(P_A|M_B,\vect{T})$ and $p(P_B|M_A,\vect{T})$ determine the weight of a given sample. The first one expresses the likelihood of $\vect{T}$ given the patches observed in $A$ and the mask observed in $B$. It essentially states that the transformation $\vect{T}$ has to be such that the patches observed in $A$ fit into the mask observed in $B$. Conversely the second term assures that the patches from $B$ fit into the mask from $A$.\\
Now we will express $p(P_A|M_B,\vect{T})$. $P_A$ is
the set of all the surface patches observed in image $A$ and can be expressed as a set of points
$\{\vect{a}_{1}, \vect{a}_{2}, ... , \vect{a}_n\}$ . Similarly we have
$P_B = \{\vect{b}_{1}, \vect{b}_{2}, ... , \vect{b}_{m}\}$. We can now write
\begin{align}p(P_B| M_A, \vect{T}) &= p(\vect{b}_{1}, \vect{b}_{2}, ... , \vect{b}_{m}| M_A,\vect{T}) \end{align}
Given the mask $M_A$ observed in image A, we look at the points $P_B$ observed in image $B$ as independent observations:
\begin{align}
\Aboxed{p(P_B| M_A, \vect{T}) &= \prod_{j = 1}^m p(\vect{b}_j| M_A,\vect{T}) \label{eq:P_B}}
\end{align}
After the derivation in Appx.~\ref{ap:der} we have\\

\hspace{-0.5cm}
\fbox{
\begin{minipage}{\columnwidth}
\begin{align}
 p(\vect{b}_j| M_A,\vect{T})  \propto &K_2\sum_{i = 1}^{n} e^{-\frac{1}{2} ([\vect{b}_j]_A - \vect{a}_{i})_{w,h}^{\top}D([\vect{b}_j]_A - \vect{a}_{i})_{w,h}}\label{eq:pbd}~~~~~~~ \\ &(1+erf(\vect{v}^{\top} ([\vect{b}_j]_A - \vect{a}_{i})))\notag
\end{align}
\end{minipage}
\hspace{-1cm}
}\\
\vspace{0.3cm}

with $D, \vect{v}, K_2, Z$ defined in Appx.~\ref{ap:der}.
The second term in Eq.~\ref{eq:sampling2}, $ p(P_A| M_B, \vect{T})$, can be expressed analogously.
\subsection{Discussion}
The first term in Eq.~\ref{eq:pbd} is a Gaussian over the parameters $w,h$ with mean $(w_i, h_i)^\top$. This term  accounts for the fact that the closer $[\vect{b}_j]_A$ is to a pixel $i$, the likelier it is that the point which has been observed at $\vect{b}_j$ in image $B$ is observed in pixel $i$ in image $A$.
The second term goes to zero if the depth of $[\vect{b}_j]_A$ is smaller than the depth at the pixel where it is projected on in image $A$, which is necessary in order to respect the mask $M_A$.\\
\begin{figure}[h]
 \centering
 \includegraphics[width = 0.45\textwidth]{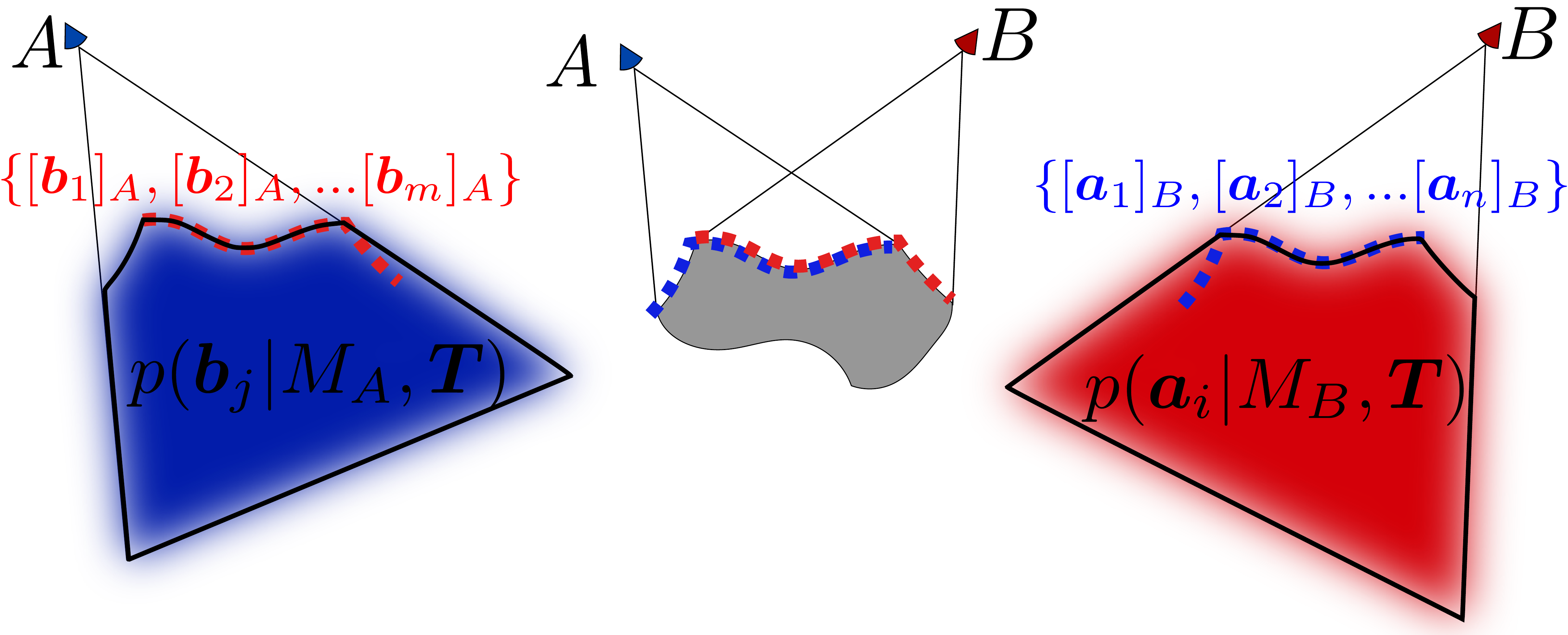}
 \caption{Schematic representation of $p(\vect{b}_j| M_A,\vect{T})$ (blue), $p(\vect{a}_i| M_B,\vect{T})$ (red), $\{\vect{a}_{1}, \vect{a}_{2}, ... , \vect{a}_n\}$ (blue dots) and $\{\vect{b}_{1}, \vect{b}_{2}, ... , \vect{b}_{m}\}$ (red dots)}
 \label{fig:principal}
\end{figure}
Given that $p(P_B| M_A, \vect{T})$ (see Eq.~\ref{eq:P_B}) is the product of all $p(\vect{b}_j| M_A,\vect{T})$, it is zero if any $p(\vect{b}_j| M_A,\vect{T})$ is zero. This result is illustrated in Fig.~\ref{fig:principal}, all of the red points have to be inside the blue area. 

\section{Implementation}
The only parameter that has to be determined for our algorithm is the covariance matrix of the camera uncertainty (Eq.~\ref{eq:emissionprob}). This is however not a parameter that has to be optimized, it represents a meaningful quantity and should be estimated for the depth camera that is used. For our experiments with the Kinect camera we estimated the covariance matrix to be isotropic with $\sigma = 0.002$, which corresponds approximately to the resolution in ray coordinates of the Kinect. These values are a very rough estimation of the properties of the Kinect, but they prove to work well in the experiments.\\
The core of the algorithm looks as follows:
\begin{itemize}
	\item For K samples
	\begin{itemize}
	\item Sample from $p(\vect{T}|N,P)$
	\item For all points in $B$
	\begin{itemize}
	\item if $p(\vect{b}_j| M_A,\vect{T})$ is zero, sample a new transform
	\item else $p(P_B| M_A, \vect{T}^{(l)}) ~*= p(\vect{b}_j| M_A,\vect{T})$
	\end{itemize}
	
	\item Do the same for points in $A$
	\end{itemize}
	\item Given all the $p(P_A| M_B, \vect{T}^{(l)})$ and $p(P_B| M_A, \vect{T}^{(l)})$ we can compute the covariance matrix and the mean of $\vect{T}$ according to Eq.~\ref{eq:sampling} and Eq.~\ref{eq:sampling2}.
\end{itemize}

\section{Results}
As mentioned in the introduction, the algorithms we want to compare against
are ICP and feature mapping. We used the implementations in the Point Cloud
Library (PCL) of these algorithms for our evaluation. We employed FPFH features
which are described in \cite{rusu09_fpfh}.\\ 
\begin{figure}[h]
 \label{fig:objects}
 \centering
 \includegraphics[width=0.45\textwidth]{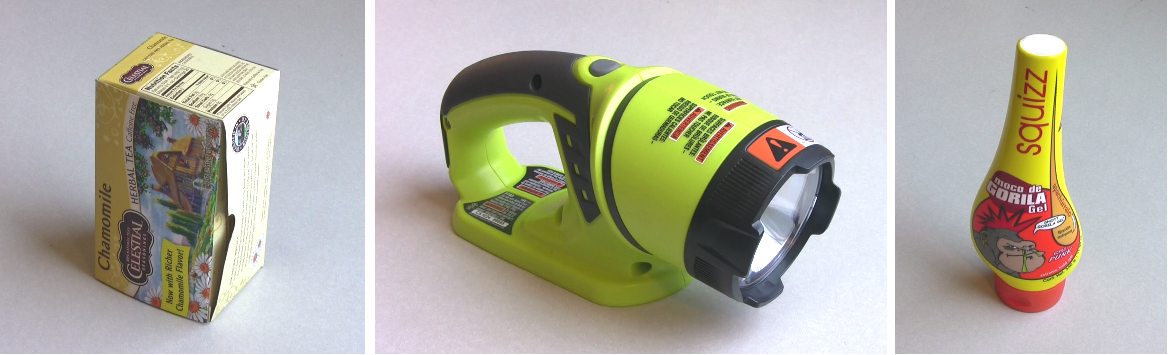}
 \caption{Box, flashlight and tube.}
\end{figure}
Our dataset consists of three objects, a box, a flashlight and a
tube. Our dataset is small, the three objects
however have a big variety in shape as seen in Fig.~\ref{fig:objects},
and therefore this evaluation gives us a reasonable idea about the
performance of our algorithm. Admittedly a broader evaluation will be
necessary for a more precise assessment of the performance. In the
associated video the algorithm is applied to a series of different
objects on a
tabletop \footnote{http://youtu.be/oWiNbItu2yM}.

 Each of the three objects has been rotated in steps of about
 $25^o$ and translated by a few $cm$ 14 times on a tabletop. At each
 step we acquire a depth image and measure the object's exact position
 and orientation which will be used as ground truth.
 For evaluation we will align each image to the next,
 which gives a total of 13 alignments per object.

We compare our algorithm to ICP, feature mapping and feature mapping with subsequent ICP. We use
the implementations of these algorithms in the Point Cloud Library
(PCL) \cite{rusu11}. The feature mapping algorithm uses Fast Point
Feature Histograms (FPFH) as shape features \cite{rusu09_fpfh}. We
used these algorithms to our best knowledge and implemented them as
suggested in tutorials of PCL. We do not claim that the performance we
measure here for ICP and feature mapping is the maximum that can be achieved with these
algorithms, but it serves as a good point of comparison for our new
algorithm.
\subsection{Evaluation of Alignment Performance without Nonvisual Information}
In order to obtain a general estimate of the alignment performance of
our algorithm we only make very general assumptions for the sampling distribution
$p(\vect{T}|N,P_A,P_B)$. We will assume that we have no information $N$, we
do thus \emph{not} use the information that object has only been
translated and rotated on a table top. We only assume that the center
of mass of $P_A$ will be no further than 4cm from the center of mass
of $P_B$ in the aligned images and that the object will not be rotated
by more than 50 degrees at a time. Note that these assumptions leave a
very big search space open, and, therefore, we have to draw a very
large amount of samples -- about 100 million -- and the algorithm is
thus slow and takes about 30 seconds per image. ICP took about 1 second
and feature mapping took about 5. In practice however we will have much stronger
sampling distributions which will accelerate our algorithm considerably.

\begin{figure}[h]
 \centering
 \includegraphics[width=0.4\textwidth]{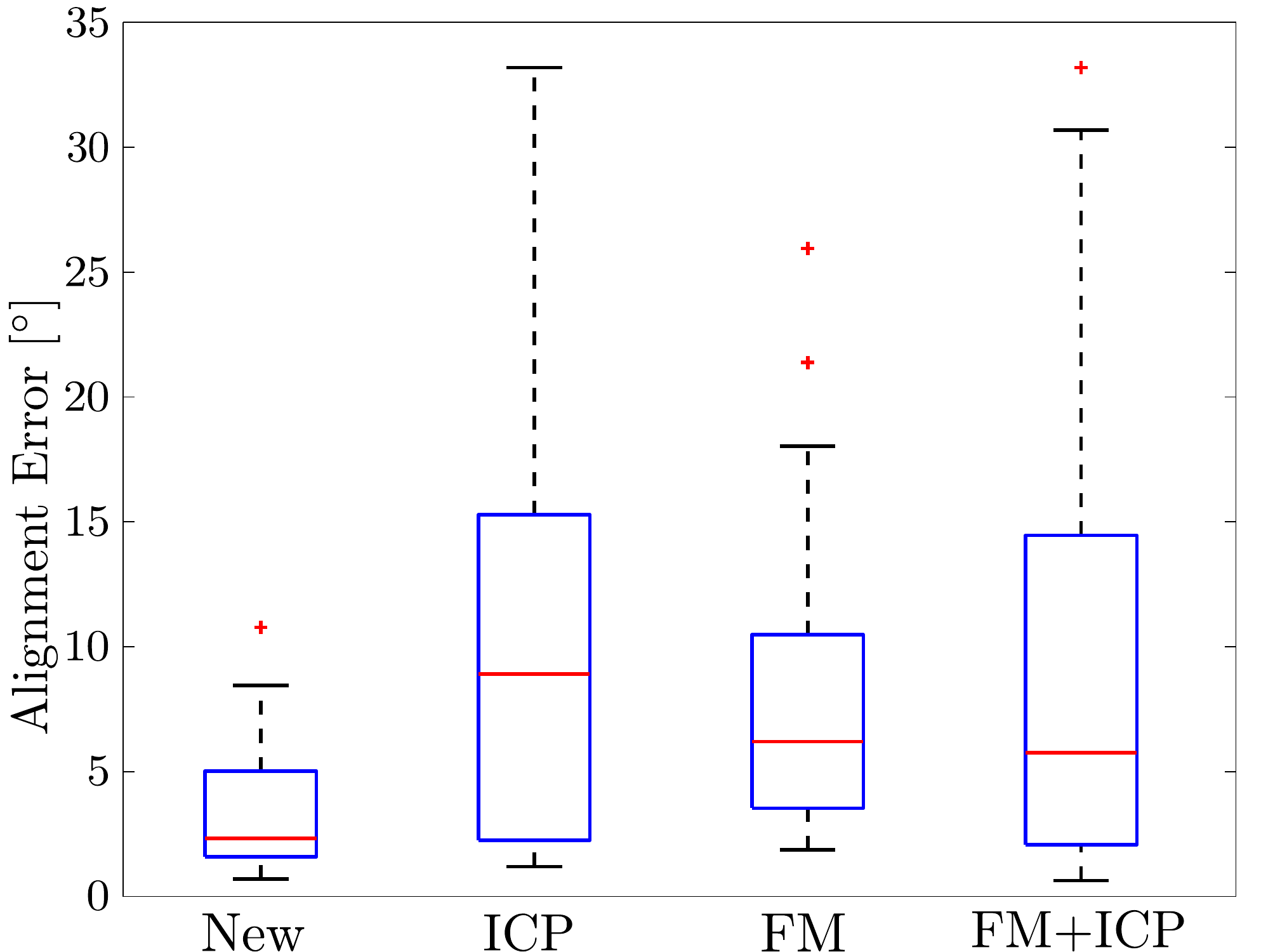}
 \caption{Boxplot of alignment error for different algorithms.}
 \label{fig:boxplot}
\end{figure}

In Fig.~\ref{fig:boxplot} we present the box-plots of the alignment error in degree
of the four algorithms for all the objects. Our algorithm performs
favorably compared to these implementations of ICP and feature mapping. We will now
try to investigate how this advantage emerges.

\begin{figure}[h]
 \centering
 \includegraphics[width=0.4\textwidth]{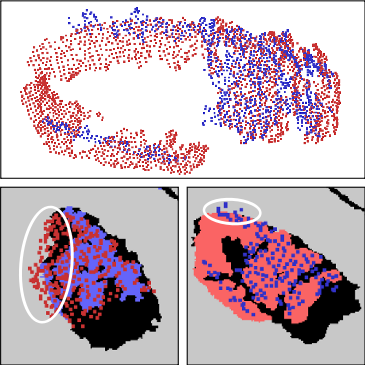}
 \caption{Flashlight aligned by ICP. Alignment error = $32^o$.}
 \label{fig:icp_flashlight}
\end{figure}
Fig.~\ref{fig:icp_flashlight} shows an alignment performed by ICP with
an error of $32^{\circ}$. The top image shows the aligned point
clouds. The two bottom images represent the information about the
mask. The left one illustrates
$p([\vect{b}_j]_A|M_A,\vect{T})$. In the blue area the object has
been observed, in the gray area background has been observed, and in
the black area no observation has been made. The red points represent
the points observed in $B$ projected into image $A$,
$[\vect{b}_j]_A$. The result of our derivation suggests that the red
points can only be in the blue or black area. If a point
$[\vect{b}_j]_A$ is located on a pixel $\vect{a}_j$ in the blue
area its distance to the camera $r$ has to be approximately equal or
larger than the depth measured at $\vect{a}_j$. If the point is
located in the black area its distance to the camera can be arbitrary
because no depth has been measured at the corresponding pixel.

ICP however only uses the information contained in the point clouds,
which are quite sparse in the considered images. Looking at the top
image of Fig.~\ref{fig:icp_flashlight} it does not surprise that ICP
performs poorly on this data. If we look at the two bottom images
however we can see that many of the projected points are in front of
the background. Taking this information into account we thus know that
this alignment is not correct.
\begin{figure}[h]
 \centering
 \includegraphics[width=0.4\textwidth]{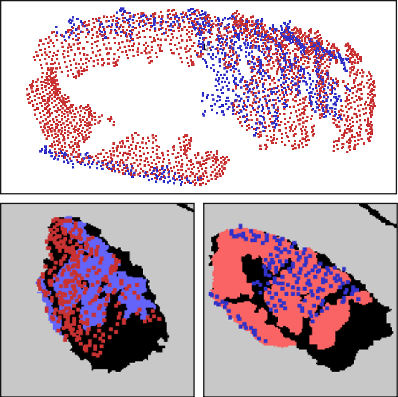}
 \caption{Flashlight aligned by new algorithm.  Alignment error = $2.1^o$.}
 \label{fig:alpha_flashlight}
\end{figure}
In Fig.~\ref{fig:alpha_flashlight} the alignment of the same two
images by our algorithm is shown. Even though the point clouds are
quite sparse it has performed well thanks to the information about the
mask.

\begin{figure}[h]
 \centering
 \includegraphics[width=0.4\textwidth]{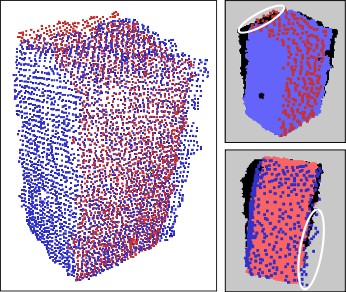}
 \caption{Box aligned by feature mapping followed by ICP.  Alignment error = $86^o$.}
 \label{fig:icp_fm_box}
\end{figure}

Fig.~\ref{fig:icp_fm_box} illustrates a problem of
a different nature that occurred with feature mapping and ICP. The problem here is
that this box, looking only at the point clouds, allows different
alignments. The red point cloud should be rotated about $90^\circ$ to
the left. The box fortunately is a little bit broader than wide.
Taking the mask into account we can thus resolve this ambiguity. In
the small image on the right on the bottom we see that many blue
points are in front of the background which enables our algorithm to
discard this alignment. Our algorithm aligned these images with an
error of $2.7^{\circ}.$

These results illustrate that taking the
mask into account can resolve important problems.

\subsection{Evaluation of Alignment Performance with Nonvisual Information}

Now we will show the benefits of taking nonvisual information $N$ about
the transformation into account. The dataset we have been working on
consists of translations and rotations on a tabletop. Before we did
not make use of this information. Now we include this information in the
sampling distribution of our algorithm. We thus only sample from translation and
rotations in the plane of the table. Of course our search space is
much smaller now, and therefore we need less samples. The
computational time is reduced to about 0.5 seconds per alignment.
\begin{figure}[h]
 \centering
 \includegraphics[width=0.4\textwidth]{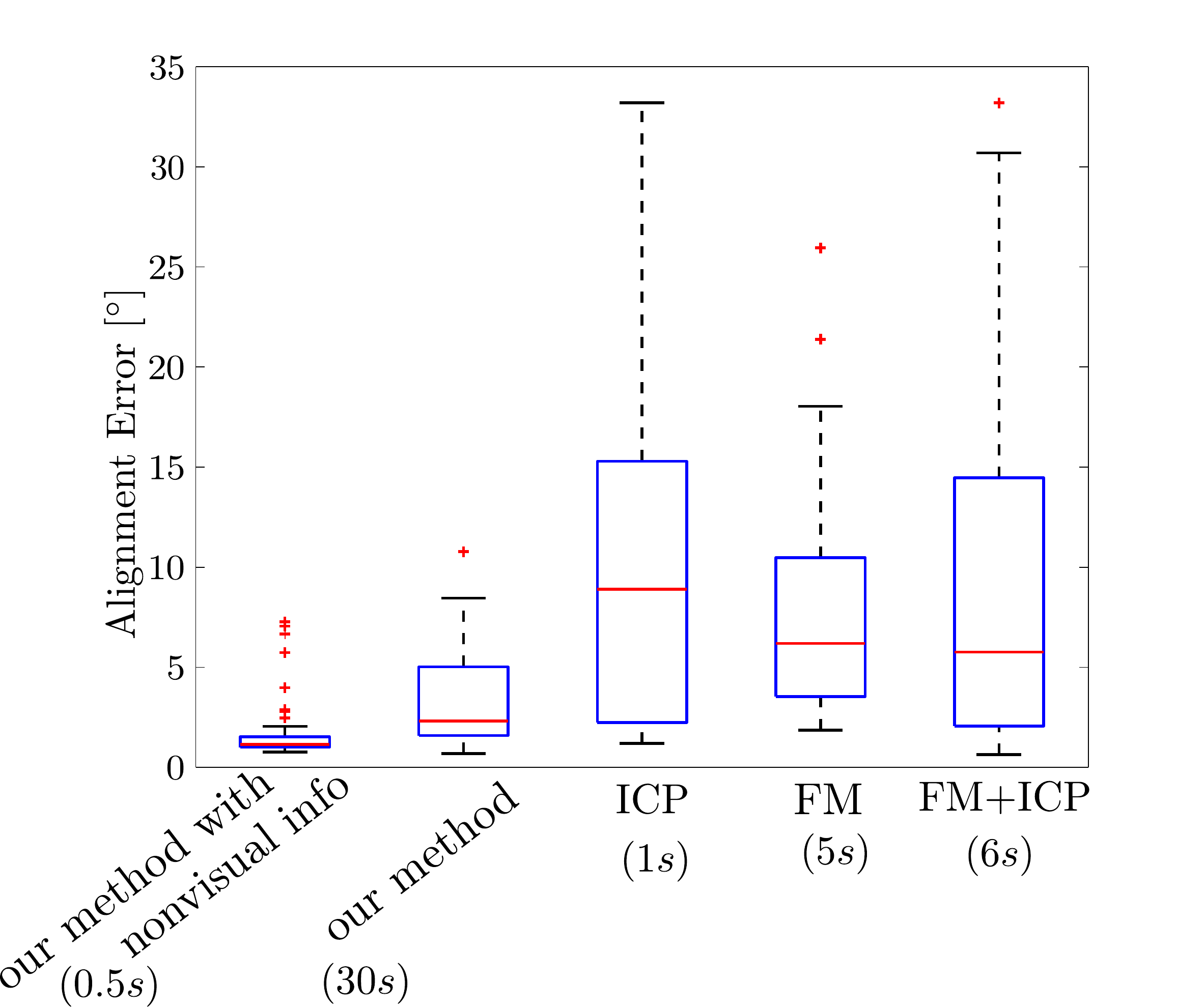}
 \caption{Boxplot of alignment error for different algorithms.}
 \label{fig:prior}
\end{figure}
The additional information of course also contributes to the alignment
performance as we can see in Fig.~\ref{fig:prior}. There may be ways
to make use of nonvisual information in ICP and feature mapping algorithms as well, it
does however not emerge naturally and we did not try to do so.
\begin{figure}[h]
 \centering
 \includegraphics[width=0.4\textwidth]{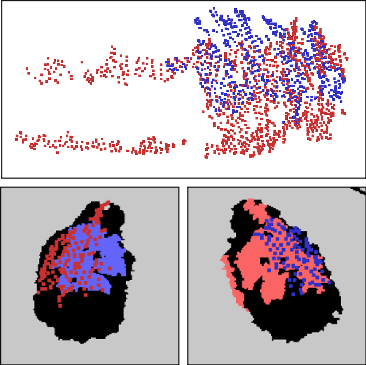}
 \caption{Very sparse point clouds of flashlight aligned with help of nonvisual information.  Alignment error = $6.3^o$.}
 \label{fig:prior}
\end{figure}
In Fig.~\ref{fig:prior} two very sparse point clouds are shown. Even
for a human it is hard to tell how these should be aligned. ICP, feature mapping
and feature mapping with subsequent ICP all aligned these point clouds with an
error of at least $51^\circ$. Our algorithm without the table prior
produced an error of $11^\circ$. With the prior however these points
are aligned with an error of only $6.3^\circ$. This example illustrates
that the combination of nonvisual information and the
information from the mask can be complementary. Even this point
cloud of very bad quality has been aligned almost correctly.

\subsection{Evaluation of Alignment Performance with Loop-closure}
\begin{figure}[h]
 \centering
 \includegraphics[width=0.45\textwidth]{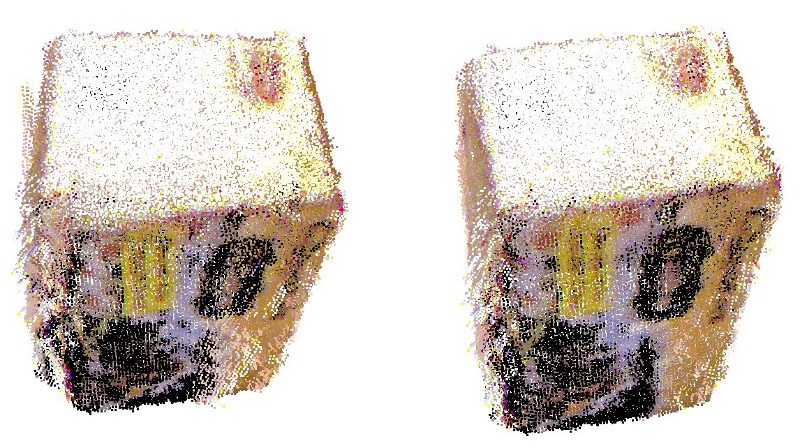}
 \caption{The transformations on the right have been optimized, on the left not.}
 \label{fig:opt}
\end{figure}

We argued that as output of the alignment we prefer a probability
distribution to a point estimate. As an example why this is useful we
will merge all the point clouds of the box, aligned by our algorithm
with the table prior, into one point cloud, as shown in
Fig.~\ref{fig:opt}. Frame 1 has been aligned to image 2, image 2 to
image 3 and so on. Between the first and the last image the object has
been rotated by around $360^\circ$, we can thus close the loop and
align the last to the first image. Therefore we now have redundant
information about the transformation of each image, and can optimize
these transformations. This optimization is problematic if we only
have a point estimate of each transformation. Fortunately however we
can compute the mean as well as the covariance matrix of each
transformation. We can thus estimate the probability of each
transformation assuming that its distribution is normal with
covariance matrix and mean as computed by our algorithm. We
numerically maximize the joint probability of all the transformations
using the graph optimization algorithm described in \cite{g2o}.
The difference between the transformations which are optimized and the ones which are not is illustrated in Fig.~\ref{fig:opt}. This illustrates one of the advantages of having a posterior distribution rather than a point estimate.

\addtolength{\textheight}{-12cm}

\section{Conclusion and future work}
The results of our evaluation are promising, but for
a full assessment of the performance of our novel algorithm many more experiments are
necessary. The derivation of this algorithm is general and does not assume in any
way that the object is on top of a table. Note that we
only used this information where we explicitly mentioned it. It might however
be favorable for the performance of our algorithm because the
depth camera always manages to measure the depth on pixels which are
on the table top. This gives us a lot of information about the
mask. The next step will be to measure the performance of the
algorithm in other cases, such as when the object is held by the
robot hand.

The core of our algorithm is sampling, it can thus
easily be parallelized or even implemented on a GPU in order to reduce
 the computational time.

In our sampling distribution
$p(\vect{T}|N,P_A,P_B)$ we have barely used the information coming from
$P_A,P_B$. This information is not very important if we already have a
good idea how the object has moved given $N$. If we have however no nonvisual information about how the object has moved
we can make assumptions based on $P_A,P_B$. These assumptions are
specific for a given application. If we know for example that we will
observe only objects that are much longer in one dimension than in the
others, then we can assume that the first Principal Component of $A$ is
approximately aligned with the first Principal Component of
$B$. Another possibility is to employ features. If we compute features
for each point in $A$ and $B$ we can create a set of possible
matches. Then we can sample from these matches, three at a time, which
gives us a sample for $\vect{T}$. We might however inherit problems of feature mapping
algorithms.

When we ran the algorithm on the robot we moved
its arms manually. This was of course only for evaluation, a possible application
 of the algorithm is to be used in the context of manipulation tasks. There are numerous  possibilities, such as using
the algorithm in a grasping pipeline. If the robot encounters, for
example, an object which does not have an obvious associated grasp
observing it from only one side, we can start poking it with actions that minimize the uncertainty in the alignment. While the
object is being moved around, our algorithm tracks it and completes a
model. The more information we gain, the more likely are we to select
the correct grasp.

In summary, we can say that there are many
applications and possible extensions for this algorithm. Its most
important feature is that due to its general formulation, it can make
use of all the information available in a given case.



\addtolength{\textheight}{30cm}

\bibliographystyle{IEEEtran}
\bibliography{paper}

\onecolumn

\appendix

\subsection{Derivation of $p(\vect{b}| M_A,\vect{T})$}\label{ap:der}
As explained in Appx.~\ref{ap:coord}, the transformation from ray coordinates in image $A$ to ray coordinates in image $B$ is not linear, it can however be approximated linearly around a point. Point $\vect{b}$ is expressed in ray coordinates in image $B$ and $\vect{a}$ is expressed in ray coordinates in image $A$.
\vspace{-0.3cm}
\begin{align}
p(\vect{b}| M_A,\vect{T})   &=\int\limits_{-\vect{\infty}}^{\vect{\infty}} p(\vect{b}|\vect{a},\vect{T})  p(\vect{a}|M_A)d\vect{a}\label{eq:pbj}
\end{align}
%
$p(\vect{b}|\vect{a},\vect{T})$ expresses the probability distribution over the position of a point $\vect{b}$ observed in $B$ given that we have observed the same point at $\vect{a}$ in $A$.  With $\vect{s}$ being the underlying point, expressed in ray coordinates in image $A$, we can write
\begin{align}
p(\vect{b}|\vect{a},\vect{T}) = \int\limits_{-\vect{\infty}}^{\vect{\infty}}  p(\vect{b}|\vect{s},\vect{T}) p(\vect{s}|\vect{a},\vect{T}) d\vect{s}%
~~~\text{and inserting Eq.~\ref{eq:emissionprob} we obtain}%
~~~ = \int\limits_{-\vect{\infty}}^{\vect{\infty}} \mathcal{N}(\vect{b}|[\vect{s}]_B, L)   \mathcal{N}(\vect{s}|\vect{a}, L) d\vect{s}
\end{align}
Given that $[\vect{s}]_B$ is only relevant in the neighborhood of $\vect{b}$ we can replace $[\vect{s}]_B$ by its linear approximation around $\vect{b}$ obtained in Appx.~\ref{ap:coord}:
\vspace{-0.3cm}
\begin{align}
& p(\vect{b}|\vect{a},\vect{T})  = \int\limits_{-\vect{\infty}}^{\vect{\infty}} \mathcal{N}(\vect{b}|\vect{b} +   Q_BRQ_A^{-1}  (\vect{s}-[\vect{b}]_A), L)   \mathcal{N}(\vect{s}|\vect{a}, L) d\vect{s}\\
\Aboxed{& p(\vect{b}|\vect{a},\vect{T})  =  K_1 e^{-\frac{1}{2}(\vect{a}-[\vect{b}]_A)^{\top}\Lambda(\vect{a}-[\vect{b}]_A)}}\\
& K_1 = \frac{1}{(2\pi)^{3/2}|L+Q_BRQ_A^{-1} L Q_A^{{-1}^{\top}}R^{\top}Q_B^{\top}|^{1/2}},~~\Lambda^{-1} =  Q_A R^{-1}Q_B^{-1}LQ_B^{\top^{-1}}RQ_A^{^{\top}} +L  
\end{align}

As explained in the assumptions section, the whole term inside the integral of Eq.~\ref{eq:pbj} can be approximated as being constant within the range of a pixel. The integral over $w$ and $h$ thus becomes a sum over the number of pixels $n$:
\begin{align}
p(\vect{b}| M_A,\vect{T})  & \propto  \sum_{i = 1}^{n} \int\limits_{-\infty}^{\infty} p(\vect{b}|w_i,h_i,r)  p(w_i,h_i,r|M_A)dr
\end{align}
$p(w_i,h_i,r|M_A)$ is the probability distribution over the observation of $\vect{b}$ in $A$, given the mask $M_A$. This probability distribution is equal to zero in the green area of Fig.~\ref{fig:concept} because we know that no part of the object has been observed there. Everywhere else it is uniform because we have no further information, considering only the mask. The green area, for a pixel $(w_i, h_i)$, corresponds to the range between the camera and the depth measured at the aforesaid pixel $(r_i)$. Therefore the probability distribution is equal to zero for $r<r_i$ and uniform for $r \geq r_i$. This can easily be translated into limits for the integral, and we obtain
\vspace{-0.3cm}
\begin{align}
p(\vect{b}| M_A,\vect{T})  & \propto  \sum_{i = 1}^{n} \int\limits_{r_i}^{\infty}   p(\vect{b}|w_i,h_i,r) dr
\end{align}%
\vspace{-0.2cm}
We can now integrate and obtain
\begin{align}
\Aboxed{p(\vect{b}| M_A,\vect{T})  &\propto K_2\sum_{i = 1}^{n} e^{-\frac{1}{2} ([\vect{b}]_A - \vect{a}_i)_{w,h}^{\top}D([\vect{b}]_A - \vect{a}_i)_{w,h}} (1+erf(\vect{v}^{\top} ([\vect{b}]_A - \vect{a}_i))}\\
\Lambda^{-1} &=  Q_A R^{-1}Q_B^{-1}LQ_B^{\vect{T}^{-1}}RQ_A^{^{\top}} +L, ~~K_2 = \frac{1}{|L+Q_BRQ_A^{-1} L Q_A^{{-1}^{\top}}R^{\top}Q_B^{\top}|^{1/2}}\\
D &= \frac{1}{\Lambda_{33}} \begin{bmatrix} \Lambda_{11}\Lambda_{33}-\Lambda_{31}^2 & \Lambda_{33}\Lambda_{21}-\Lambda_{31}\Lambda_{32}\\
                             \Lambda_{33}\Lambda_{21}-\Lambda_{31}\Lambda_{32} & \Lambda_{22}\Lambda_{33}-\Lambda_{32}^2\end{bmatrix}, ~~
\vect{v}  = \frac{1}{\sqrt{2 \Lambda_{33}}} \begin{bmatrix} 
                                              \Lambda_{31}\\ \Lambda_{32}\\ \Lambda_{33}
                                             \end{bmatrix}
\end{align}

\subsection{Linear approximation to ray coordinate transformation}\label{ap:coord}
We want to linearly approximate the transformation from ray coordinates in image $A$, to ray coordinates in image $B$, around the point $\vect{b}$, which is defined in ray coordinates in image $B$. With $\vect{s}$ defined in image $A$, it is straightforward to show that
\begin{align}
\Aboxed{[\vect{s}]_B  &\approx \vect{b} +   Q_B R Q_A^{-1}  (\vect{s}-[\vect{b}]_A)} %
~~\text{with} ~~Q_A = \begin{bmatrix} \frac{\partial w}{\partial x}&\frac{\partial w}{\partial y}&\frac{\partial w}{\partial z}\\
							\frac{\partial h}{\partial x}&\frac{\partial h}{\partial y}&\frac{\partial h}{\partial z}\\ 
							\frac{\partial r}{\partial x}&\frac{\partial r}{\partial y}&\frac{\partial r}{\partial z}\\ \end{bmatrix}_{[\vect{b}]_{A} } ~\text{and}~~ Q_B = \begin{bmatrix} \frac{\partial w}{\partial x}&\frac{\partial w}{\partial y}&\frac{\partial w}{\partial z}\\
							\frac{\partial h}{\partial x}&\frac{\partial h}{\partial y}&\frac{\partial h}{\partial z}\\ 
							\frac{\partial r}{\partial x}&\frac{\partial r}{\partial y}&\frac{\partial r}{\partial z}\\ \end{bmatrix}_{\vect{b} }
\end{align}

\end{document}